\documentclass[letterpaper, 10 pt, conference]{ieeeconf}  
\IEEEoverridecommandlockouts                    

\usepackage[usenames,dvipsnames]{xcolor}

\makeatletter

\let\proof\@undefined
\let\endproof\@undefined
\makeatother

\usepackage{amsmath} 
\usepackage{amssymb}  
\usepackage{amsthm}
\usepackage{amsfonts}
\usepackage{mathtools}
\usepackage{verbatim}

\usepackage[outdir=./epsToPdf]{epstopdf}

\usepackage{graphicx}
\usepackage{caption}
\usepackage{subcaption}
\usepackage{epstopdf}

\usepackage{soul} 

\usepackage{times}

\usepackage[us]{datetime}
\usepackage{caption}
\usepackage{todonotes}

\usepackage{multicol}

\usepackage{latexsym}

\usepackage{cite}

\title{\LARGE \bf Does spontaneous motion lead to intuitive Body-Machine Interfaces? \\ A  fitness study of different body segments for wearable telerobotics}

\author{
Matteo Macchini, \textit{Student Member, IEEE}, Jan Frogg, Fabrizio Schiano, \textit{Member, IEEE}, \\ and Dario Floreano, \textit{Senior Member, IEEE}
\thanks{The authors are with the Laboratory of Intelligent Systems, École Polytechnique Fédérale de Lausanne, CH-1015 Lausanne (EPFL), Switzerland.}%
}

\newcommand{\kron}{\otimes}

\newcommand{\matr}[1]{\boldsymbol{#1}}

\newcommand{\zeros}[2]{
\ifthenelse{\equal{#2}{1}}{\vect{0}_{#1}}{\matr{\cancel{O}}_{#1 \times #2}}
}
\newcommand{\ones}[2]{
\ifthenelse{\equal{#2}{1}}{\vect{1}_{#1}}{\matr{1}_{#1 \kron #2}}
}

\newcounter{simulationcase}

\definecolor{mygray}{gray}{0.75} 
\newcommand{\old}[1]{}  

\newcommand{\reffig}[1]{Fig. \ref{#1}}

\usepackage[dvipsnames]{xcolor}
\colorlet{orange_maxim}{green!10!orange!90!}

\begin{document}

\maketitle
\thispagestyle{empty}
\pagestyle{empty}

\begin{abstract}
Human-Robot Interfaces (HRIs) represent a crucial component in telerobotic systems.
Body-Machine Interfaces (BoMIs) based on body motion can feel more intuitive than standard HRIs for naive users as they leverage humans' natural control capability over their movements.
Among the different methods used to map human gestures into robot commands, data-driven approaches select a set of body segments and transform their motion into commands for the robot based on the users' spontaneous motion patterns.
Despite being a versatile and generic method, there is no scientific evidence that implementing an interface based on spontaneous motion maximizes its effectiveness.
In this study, we compare a set of BoMIs based on different body segments to investigate this aspect.
We evaluate the interfaces in a teleoperation task of a fixed-wing drone and observe users' performance and feedback.
To this aim, we use a framework that allows a user to control the drone with a single Inertial Measurement Unit (IMU) and without prior instructions.
We show through a user study that selecting the body segment for a BoMI based on spontaneous motion can lead to sub-optimal performance.
Based on our findings, we suggest additional metrics based on biomechanical and behavioral factors that might improve data-driven methods for the design of HRIs.
\end{abstract}

\vspace{0.5cm}



\section{Introduction}\label{sec:introduction}


Several robotic applications, such as search-and-rescue missions \cite{casper_human-robot_2003, bruggemann_coupled_2015}, exploration of complex environments \cite{diftler_robonaut_2011, khatib_ocean_2016}, and robotic surgery \cite{taylor_medical_2003, morelli_da_2016}, require human teleoperation and intuitive Human-Robot Interfaces (HRI). 
An HRI is a system that acquires inputs from the human operator and translates them into commands for the robot. 
One of the main objectives of effective HRIs is \textit{intuitiveness}.
Despite the extensive use of the term, a formal definition of intuitiveness is missing. This term is often referred to as a compound of objective dimensions like robustness and performance and subjective human factors such as the sense of presence, transparency, and individual preferences~\cite{passenberg_survey_2010}.
In summary, we consider an interface to be intuitive if new users can master it in a short time, and if their experience is positive.

Widely used HRIs, such as remote controllers, require substantial cognitive effort and long training sessions to be mastered \cite{casper_human-robot_2003, nagatani_emergency_2013, chen_supervisory_2011}.
Moreover, HRI design relies on predefined gestures with few degrees of freedom instead of leveraging the natural human dexterity.
Recent studies have shown that using spontaneous body motion as control inputs can improve teleoperation efficiency and reduce cognitive workload~\cite{miehlbradt_data-driven_2018, rognon_flyjacket:_2018, macchini_personalized_2020}. HRIs that use body motion signals are called Body-Machine Interfaces (BoMIs).

A major challenge in the BoMI design is the definition of a \textit{mapping} function.
The mapping function transforms a set of body signals into robot commands and carries out two main tasks \cite{casadio_body-machine_2012}.
First, it selects a subset of the significant signals to control the robot (e.g., the motion of one or more body segments).
This step is known as \textit{feature selection}.
Second, it transforms those signals into robot commands employing regression or classification methods.
In this article, we experimentally assess the effectiveness of feature selection methods and challenge the assumption that spontaneous body motion always results in better HRIs.

BoMI feature-selection methods can be grouped in two categories: \textit{data-independent} and \textit{data-driven}.
Data-independent methods refer to approaches that do not leverage prior knowledge about the operators' preferred motion to control the robot.
They can be implemented by considering kinematic or functional correspondences between the human body's morphology and the robot's morphology and establishing a linear mapping between the two.
For example, the control of NASA's humanoid robot Robonaut is performed by mapping directly the operator's arm motion into commands for the robot arm \cite{ambrose_robonaut_2000}.
It is also possible to design the mapping heuristically, based on familiar and straightforward body motion, or based on the task requirements.
For example, allowing users to control a quadrotor's position by scaling the user's hand position \cite{macchini_hand-worn_2020}, or indicating the desired direction \cite{pfeil_exploring_2013, sanna_kinect-based_2013, ikeuchi_kinecdrone:_2014}, or again to control manipulators through feet motion leaving the user's hands free \cite{sanchez_four-arm_2019}.
Data-independent methods depend on both the robot's morphology and the task, and thus cannot be generalized.

Instead, data-driven methods aim at extracting the human-robot mapping from the \textit{spontaneous} motion of the operator attempting to control a robot. 
Spontaneity, in this case, refers to body motion patterns that a person would adopt to control the robot without prior instructions.
Operators are presented with actions or maneuvers performed by the robot and are asked to move their body as if they were controlling the robot with their body motion.
Data-driven approaches can be transferred across different robots as they do not depend on the kinematic model of the robot, such as robotic arms \cite{khurshid_data-driven_2015}, fixed-wing drones \cite{miehlbradt_data-driven_2018}, quadrotors \cite{cauchard_drone_2015} and wheeled robots \cite{melidis_intuitive_2018}.
In previous work on winged drones, the authors observed that most people spontaneously move their torso to control the robot \cite{miehlbradt_data-driven_2018, macchini_personalized_2020}.

In our previous work, we developed a framework, based on a data-driven approach, to automatically implement HRI mappings personalized to each individual \cite{macchini_personalized_2020}. Since human subjects display different preferred motions, we found that the users' performance can improve if the HRI is adapted to the user. However, to track the body motion of the participants, we used a motion capture system. This infrastructure is bulky, expensive, and cannot be easily deployed outside of laboratory settings. For this reason, in this work, we extended our framework to track the orientation of a single body segment with a single IMU.

Despite the efforts in designing new BoMIs, there is still no scientific evidence that feature selection based on spontaneous motion results in more intuitive HRIs. In this article, we fill this gap for the example of fixed-wing drone teleoperation. 
To reduce the scope of our work we selected a set of body segments from the upper body and let a set of participants (N=18) teleoperate the drone using these segments. 
Specifically, we selected the body segments in the upper body having 3 degrees of freedom: torso, hand, and upper arm (\reffig{f:protocol}A).
We believe that this set is representative of the task at hand. Indeed, previous work found that the torso is the most spontaneous segment for drone teleoperation~\cite{miehlbradt_data-driven_2018, macchini_hand-worn_2020}. Hands are often used for manipulation and interfacing with external devices. Finally, the upper arm serves as a baseline condition.
As there are different ways to use a body segment to interface with a robot, we employed our framework to allow each participant to use their preferred strategy.
We compared the users' performance during the drone teleoperation task and we also gathered their feedback about the teleoperation experience through a questionnaire.  
We found that motion patterns spontaneously adopted by users do not necessarily translate into better interfaces.
Finally, through a study of the participants' motion, we suggest alternative metrics based on biomechanical and behavioral factors that might improve the design process of novel HRIs based on data-driven methods.

The paper is organized as follows.
Section~\ref{sec:methods} presents an overview of the overall methodology. 
Section~\ref{sec:results} presents the results, whereas Section~\ref{sec:discussion} discusses them. Finally, Section~\ref{sec:conclusion} concludes the paper.

\section{Methods}\label{sec:methods}

\begin{figure*}[h!]
\includegraphics[width=\textwidth]{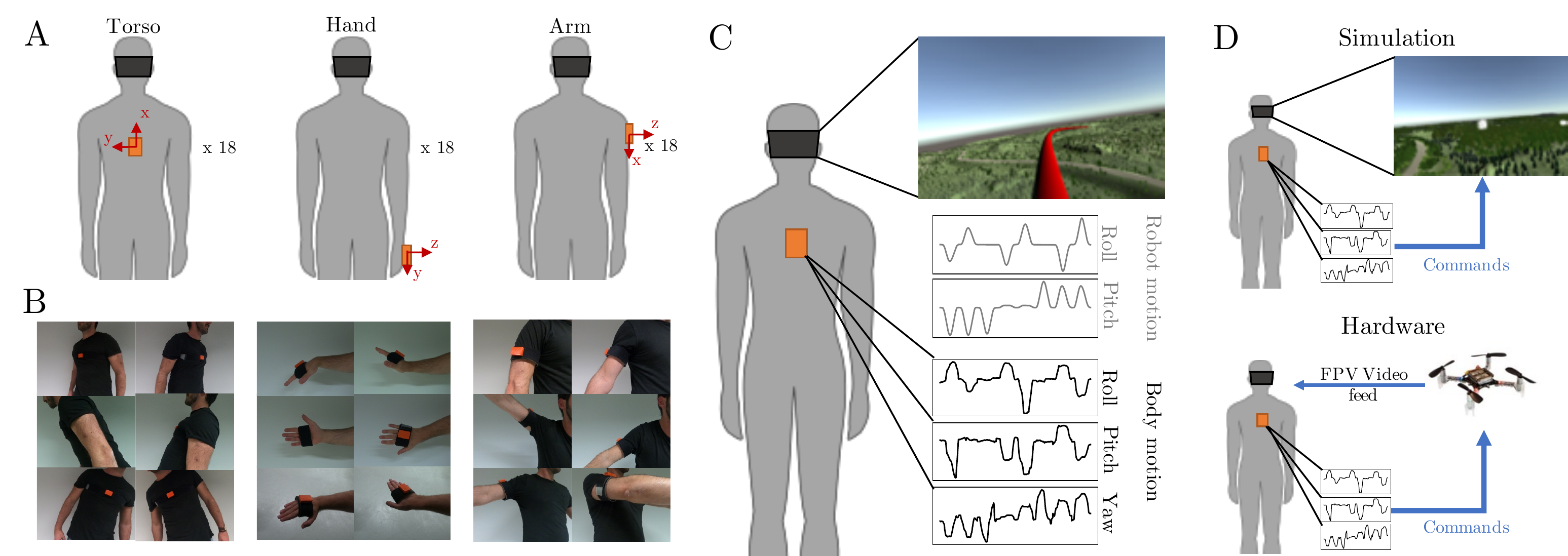}
	\caption{
     Overview of the experimental protocol. 
    (A) Anchoring point and reference frame for the three conditions corresponding to the tracked body segments. The IMU was always installed on the dominant hand and the corresponding arm.
    (B) Acquisition procedure for mobility and speed evaluation of the three body segments.
    (C) Calibration procedure. The participant observes the simulation and moves the selected body segment as if they were in control of the fixed-wing drone (here, Torso interface).
    (D) Teleoperation task. The participant controls the drone trajectory with their body using the personalized body-machine interface created during the calibration (here, simulation and hardware tests for the Torso interface).
    }

	\label{f:protocol}
\end{figure*}


\textbf{Participants: } 18 participants between the ages of 19 and 31 were recruited for our study, with the following characteristics: age = $23 \pm 2$, 14 were male and 4 were female, 16 were right-handed and 2 were left-handed.
At the beginning of each session, we explained the experimental procedure and obtained informed consent\footnote{The experiments were approved by the Human Research Ethics Committee of the École Polytechnique Fédérale de Lausanne (EPFL)}.

\textbf{Simulator:} 
we used a fixed-wing drone simulator, developed in Unity3D, to run our experiment.
The simulation displayed the trajectory of a fixed-wing drone flying at a constant speed of $12~m/s$.
The drone was modeled to mimic a realistic UAV, with three PI controllers for roll, pitch, and speed setpoints.
The drone performed a set of predefined maneuvers: two roll maneuvers (left/right), two pitch maneuvers (up/down), and four additional maneuvers consisting of a combination of pitch and roll.
The duration of each maneuver was $8s$.

\textbf{Motion data acquisition:} we captured the user's body motion with a single IMU placed on the body segment of interest.
The IMU orientation was acquired at a frequency $f = 100Hz$, in quaternion format, and converted into Euler Angles according to the ZYX convention (\reffig{f:protocol}A).

\textbf{Body segment dexterity estimation: } in the first stage, we recorded each participant's mobility and speed of motion using a fixed set of movements involving the three considered body segments.
We installed the IMU on the three body parts and acquired data, in order, for the torso, the hand, and the upper arm.
First, we evaluated the segment mobility.
We asked the participants to move the body segment on which the IMU was situated to the limit of their motion capability in the order roll-pitch-yaw in the IMU frame.
For the torso, it corresponded to the order left/right rotation - extension/flexion - right/left side bending.
For the hand, it corresponded to the order flexion/extension - pronation/supination - ulnar/radial deviation.
For the upper arm, it corresponded to the order positive/negative circumduction - abduction/adduction - flexion-extension (\reffig{f:protocol}B).
Each movement was repeated five times.
Subsequently, we measured the maximum angular speed of each body segment. We asked our participants to repeat the three movements above 
five additional times, this time moving as fast as possible.

\textbf{Personalized HRI implementation: } in the second stage, we generated a personalized interface for each participant to allow them to teleoperate a fixed-wing drone.
We installed the IMU on the user's body corresponding to one of the three tracked body segments.
The participant was asked to imitate the motion of a simulated drone with their body, by moving the segment of their upper body which was currently being tracked (\reffig{f:protocol}C).
The mapping is personalized as, even if the body part was predefined, the users were given no instructions about how to move to control the robot.
The participants wore a Head-Mounted Display (HMD) for VR during this phase, with a First-Person View (FPV) viewpoint, to increase their immersion in the virtual environment.
We acquired data synchronously from the simulator and the IMU, and the framework provided a mapping function between the IMU motion and the drone input commands.
The algorithm is based on the  method we proposed in our previous work \cite{macchini_personalized_2020}, and simplified into two main steps: data preprocessing and regression.
Moreover, it was extended to accept IMU inputs, and a filtering stage consisting of a fifth-order digital Butterworth filter was included to cope with the higher noise of the sensor.

\textbf{Teleoperation task: } after the generation of the personalized mapping for each participant, they were asked to teleoperate the drone through a path consisting of 42 waypoints (\reffig{f:protocol}D).
Each waypoint was placed at a fixed distance from the previous one, with a fixed horizontal or vertical displacement, corresponding to one of the four basic maneuvers left/right/up/down. 
Each path was pseudo-randomly generated at the beginning of each run to contain all the maneuvers in the same quantity.
The participants were instructed to steer the simulated drone as close as possible to the center of each waypoint and performed the teleoperation task twice with each body segment.
At the end of the two runs, we installed the IMU on a different body segment and repeated the calibration procedure.
To minimize bias effects, we generated pseudo-randomly the order of the three body segments to be tracked. 
Before initiating the new calibration procedure, the subjects performed a washout task \cite{khurshid_data-driven_2015}.
The goal of the task was to prevent users from getting used to the simulator dynamics and to compensate for possible learning effects. During the washout task, we inverted the signs of the commands for the drone.
Additionally, the gains of the internal PID regulators were randomly modified in a range of $\pm10\%$ from their nominal values to change the system dynamics. We did not record data during the washout tasks, and the users did not know its purpose during the experiment.

\textbf{Surveys: } 
at the end of the experimental procedure, the participants filled an additional personal feedback questionnaire comprising four questions about the teleoperation experience, listed in Table~\ref{t:quest}.

\begin{table}[th]
\normalsize
\begin{center}
\begin{tabular}{ r  l}
 \textbf{ID} & \textbf{Question} \\
 \hline
  Q1 & Which body part was the easiest to use? \\
  Q2 & Which body part was the most comfortable to use? \\
  Q3 & Which body part did you prefer using? \\
  Q4 & Why did you prefer this body part? \\
 \hline
\end{tabular}
\caption{User personal feedback questionnaire.}
\label{t:quest}
\end{center}
\end{table}
\section{Results}\label{sec:results}

In this section, we report the experimental protocols and results obtained through our study.
First, we performed an extensive user study in simulation, and later a qualitative study on a real robot. 
Our main results consist of a statistical analysis of the effects of using a specific body segment for the control of the robot.
Our study is developed on three experimental conditions, corresponding to the three groups using the aforementioned body segments:
\begin{itemize}
    \item Torso body segment users (hereafter ``Torso group")
    \item Hand body segment users (hereafter ``Hand group")
    \item Upper arm body segment users (hereafter ``Arm group")
\end{itemize}

All of our results employ the Kruskal-Wallis T-test to assess the statistical significance relative to the equality of the medians, and the Levene T-test for the equality of the variances \cite{kruskal_use_1952, brown_robust_1974}.

\begin{figure}[th]
\begin{center}
  \includegraphics[width=\columnwidth]{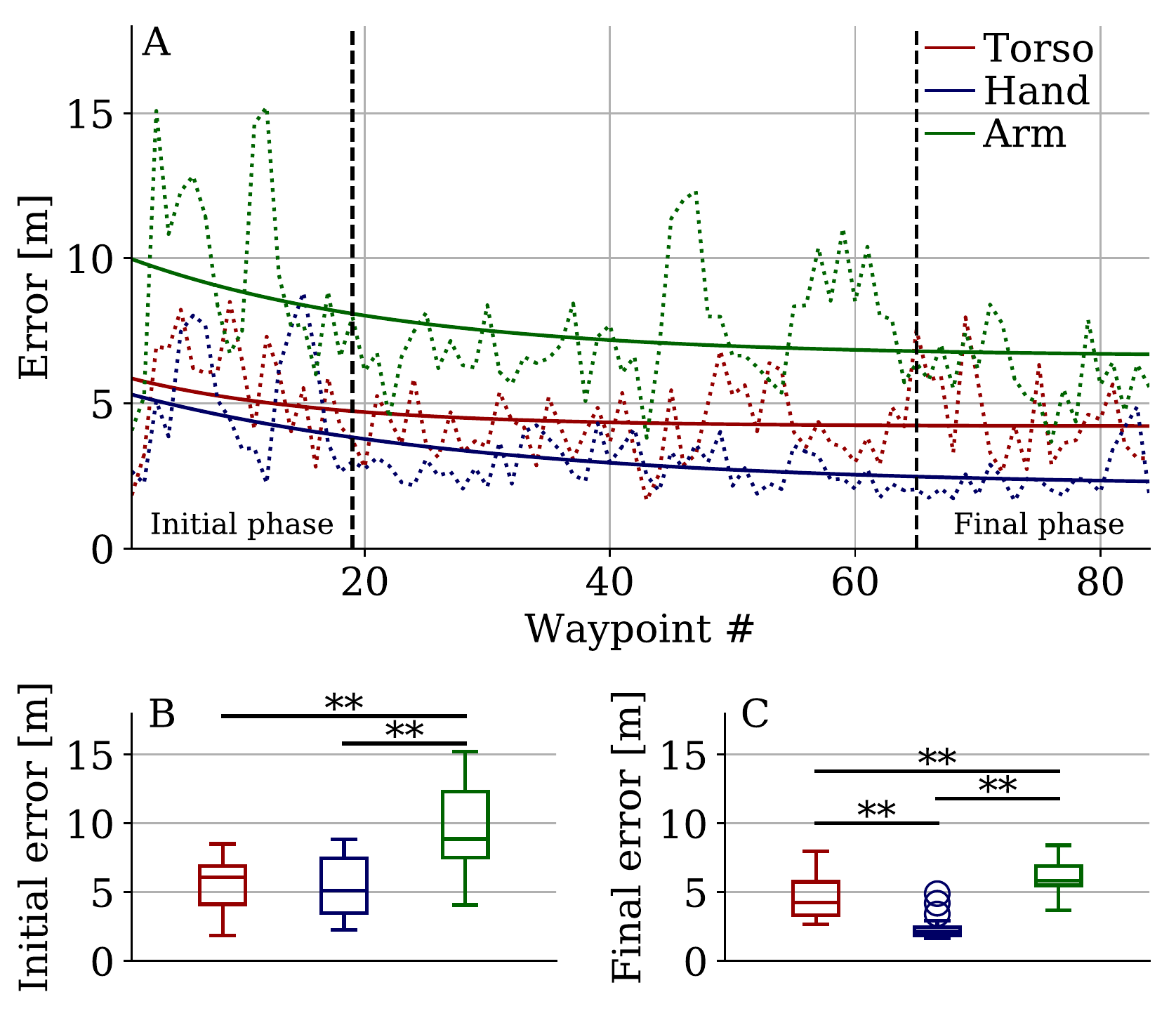}
\caption{Teleoperation error while crossing the waypoints. (A) Evolution of error during the two repetitions of the task (colored dotted lines  and exponential fit (solid line). (B) Initial error, measured in the first minute of teleoperation. (C) Final error, measured in the last minute of teleoperation. ($^{**}p<0.01$, $^{*}p<0.05$)}
\label{f:perf}
\end{center}
\end{figure}

\textit{\textbf{Hand-based interface leads to the best performance, arm to the worst:}}
first, we evaluated the participants' performance during the teleoperation.
In this phase of the experiment, we defined a personalized mapping between each participant and the simulated drone, as described in Section \ref{sec:methods}.
During the teleoperation task, the user was asked to follow a randomly generated pathway consisting of 42 waypoints, for a total of two runs.
We used the distance from the center of the waypoints as a performance metric - the shorter the distance, the better the performance.
The average across all 18 participants of this error throughout all 84 waypoints shows clear differences across the experimental groups (\reffig{f:perf}A). 
For each group, we fit a decreasing exponential curve to show the trend.
In terms of average error, the arm proved to be the worst body part for the control of the drone, whereas the hand was the best.
In all cases, the final error was lower than the initial one, proving that all the groups showed some learning effects in the task.
We considered two metrics to quantitatively evaluate the participants' performance.
First, we considered the initial error, computed as the average of the errors during the first minute of navigation corresponding, on average, to the first 18 waypoints.
Hand group ($e_H^I = 5.22 \pm 2.12 ~m$) and Torso group ($e_T^I = 5.66 \pm 1.18 ~m$) outperformed Arm group ($e_A^I = 9.64 \pm 3.38 ~m$, $ p < 0.01$) in the initial phase.
Secondly, we defined the final error, computed analogously as the average of the errors during the first minute of navigation.
At the end of the two runs, the initial differences were preserved, with Hand group ($e_H^F = 2.40 \pm 0.79 ~m$) and Torso group ($e_T^F = 4.52 \pm 1.50 ~m$) showing smaller errors than Arm group ($e_A^F = 6.11 \pm 1.20 ~m$, $p < 0.01$).
Moreover, Hand group exhibited a higher learning effect, performing significantly better than Torso group ($p < 0.01$) in the final phase.

\begin{figure}[th]
\begin{center}
  \includegraphics[width=\columnwidth]{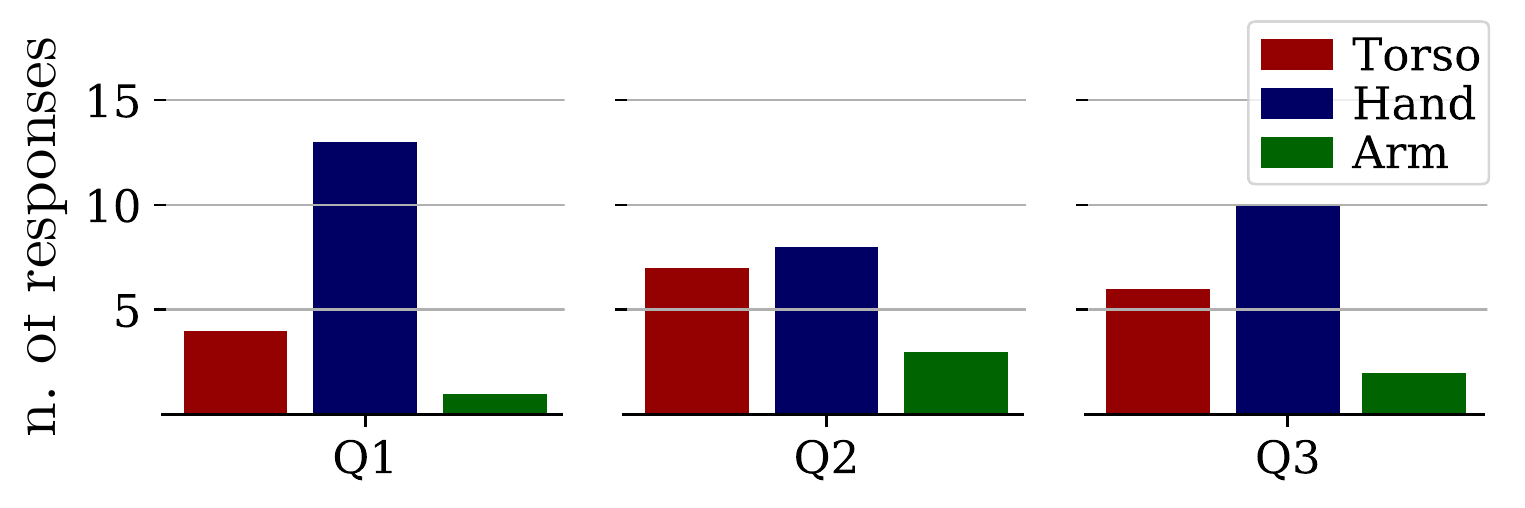}
\caption{Survey  results  relative  to  the user's preferred body segment (questions in Table \ref{t:quest}).}
\label{f:quest}
\end{center}
\end{figure}

\textit{\textbf{Hand motion is easier and preferred by users according to questionnaire results: }}
the personal feedback questionnaire responses provide insights about the users' preferred body parts to be used in the proposed task (\reffig{f:quest}).
The question on control easiness Q1 showed a clear preference towards the use of the hand.
Out of the 18 total participants, 13 reported the hand to be the easiest body segment to use, 4 chose the torso and only 1 chose the arm.
In terms of comfort (question Q2), the hand segment was again the most popular choice with 8 subjects, closely followed by the torso. The arm was identified as the most comfortable segment by 3 subjects only.
Finally, the question about the preferred body segment Q3 provided a summary of what we observed about easiness and comfort.
10 participants preferred the hand and 6 preferred the torso, with a minority of 2 participants selecting the arm as their preferred body part to use.
These results are a good indicator of the performance we observed in the teleoperation task.
Most subjects who preferred to use the hand stated in Q4 that their choice was due to a higher sensitivity, easiness of use, and precision.
The 6 people who preferred the torso argued that it felt more natural, comfortable, immersive, and it provided a more realistic experience.

\begin{figure}[th]
\begin{center}
  \includegraphics[width=\columnwidth]{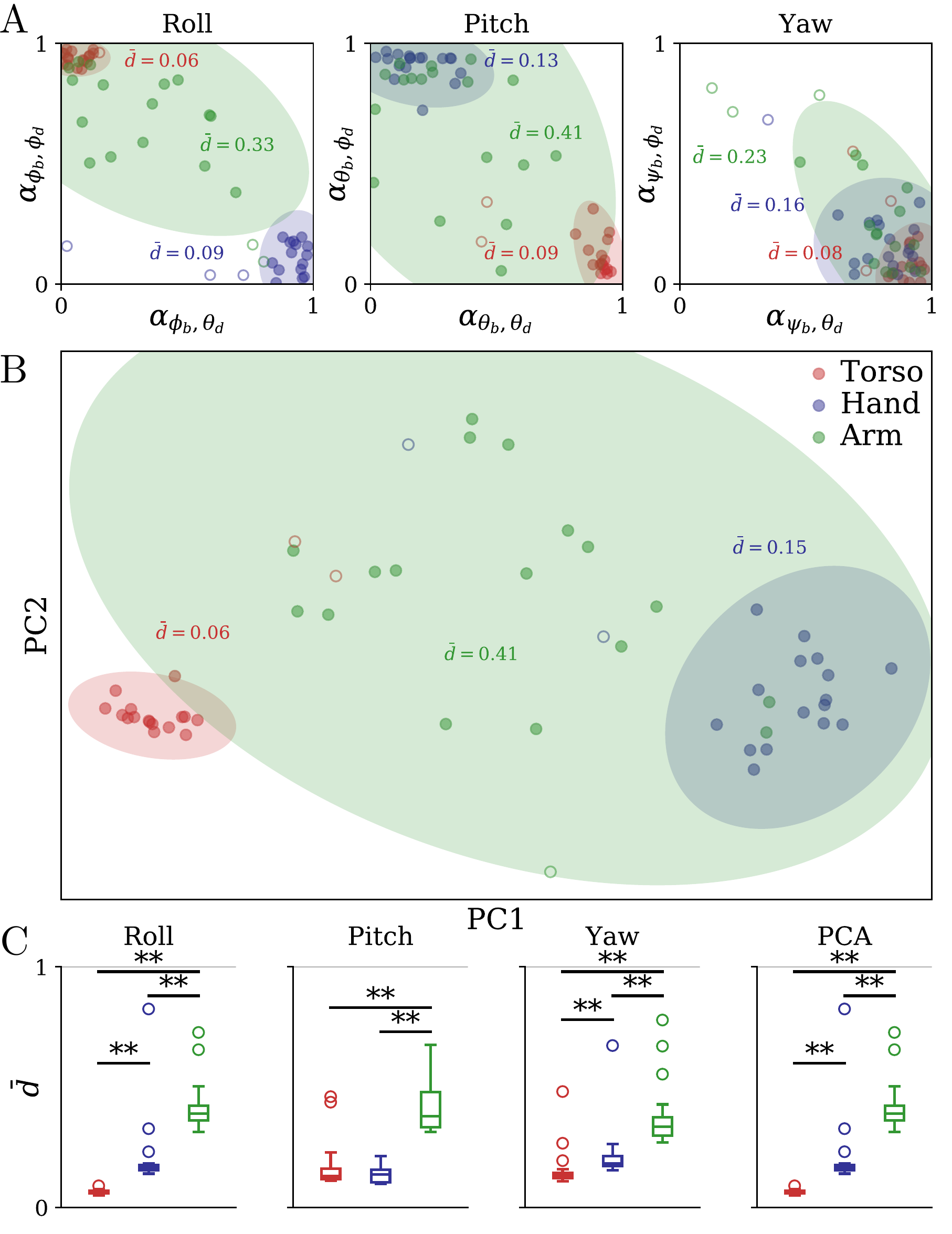}
\caption{Intra-groups agreement in terms of correlation between body segments motion and drone motion. Ellipses show three standard deviations of a bivariate Gaussian fit. (A) Pearson $\alpha$ coefficient for the three Euler angles, for drone roll (x axis) and drone pitch (y axis). (B)  PCA compression of the three datasets, confirming the higher motion variability in Arm group. (C) Average Euclidean distance between data points ($^{**} p<0.01$, $^* p<0.05$).}
\label{f:motion}
\end{center}
\end{figure}

\textbf{\textit{Arm group showed more variable motion: }} we analyzed the subjects' movements during the calibration phase and evaluated analogies and differences in their motion.
We quantified how the Euler angles of the IMU correlate with the angles of the simulated drone using the Pearson correlation coefficient $\alpha$.
For each body part, we computed six alpha values.
Let $\phi_{b}$, $\theta_{b}$, $\psi_{b}$ the body roll, pitch and yaw and $\phi_{d}$, $\theta_{d}$ the drone roll and pitch. The correlation between the quantity $X_b$ and $Y_d$ is:
\begin{equation}
    \begin{aligned}
    	&\alpha_{X_b, Y_d} = \frac{cov(X_b, Y_d)}{\sigma_{X_b} \sigma_{Y_d}}
	\end{aligned}
\end{equation}
where $ X \in \{\phi, \theta, \psi\}, Y \in \{\phi, \theta\}$. The coefficient $\alpha$ is an indicator of how a subject decided to control a specific degree of freedom of the robot.
If, for example, a high $\alpha_{\psi_b,\psi_d}$ is observed between the drone pitch $\psi_d$ and the torso flexion $\psi_b$, the user imitated the drone pitch by flexing their torso. 

 The correlation of the different features varies significantly across conditions as a representation of the participants' instinctive motion patterns (\reffig{f:motion}A).
Torso group highly employed torso rotation and bending to control the roll, and flexion/extension to control the pitch.
Similarly, Hand group controlled the drone roll using hand pronation and deviation, and its pitch using flexion/extension.
Visualizing the correlation in 2D, these common patterns appear like closely aggregated clusters with a short Euclidean distance between data points ($\bar{d}$ in \reffig{f:motion}A).
Contrarily, for Arm group we could not identify a clear pattern, as a lower intra-group agreement was observed.
The projection of the three datasets on a 2-dimensional manifold using Principal Component Analysis (PCA) confirmed the presence of two compact clusters for Torso group and Hand group, and a larger one for Arm group (\reffig{f:motion}B).
We found a significantly higher distance in the principal components data points for Arm group ($\bar{d}_{PC_{A}} = 0.41 \pm 0.075$) than both Hand ($\bar{d}_{PC_{H}} = 0.15 \pm 0.03$, $p<0.01$), and Torso groups ($\bar{d}_{PC_{T}} = 0.06 \pm 0.075$, $p_{AT}, p_{HT}<0.01$) (\reffig{f:motion}B,C).


\begin{figure}[th]
\begin{center}
\includegraphics[width=\columnwidth]{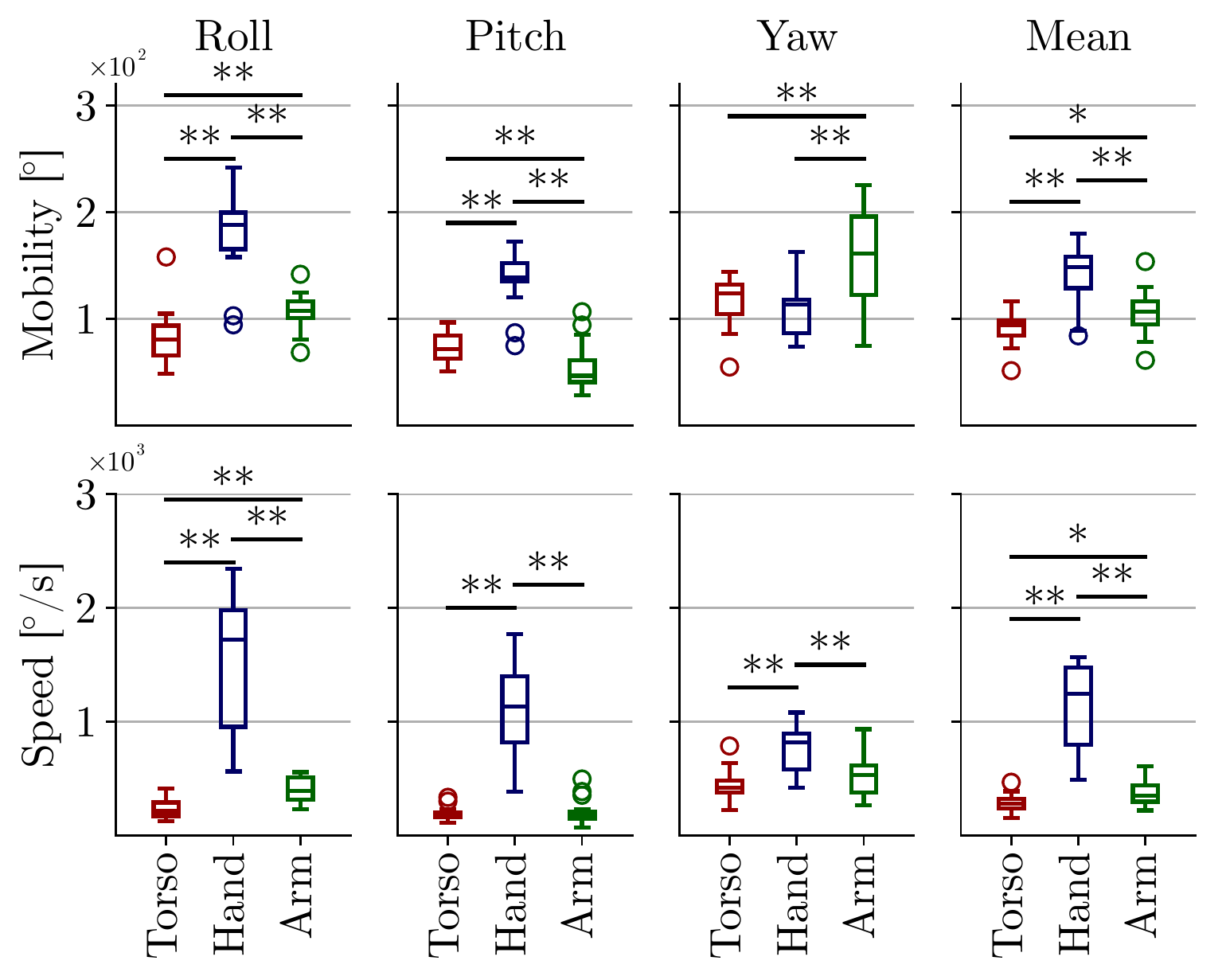}
\caption{Mobility and speed of motion of the three body segments over their three degrees of freedom and average value. The hand shows a higher dexterity compared to arm and torso. ($^{**}p<0.01$, $^{*}p<0.05$)}
\label{f:range}
\end{center}
\end{figure}

\textit{\textbf{Hand has the highest dexterity, torso has the lowest:}} 
the mobility and speed of motion acquisition resulted in large differences between the considered body parts (\reffig{f:range}).


For the mobility, we considered the difference between the average maximum angle and average minimum angle reached during the five repetitions of the movements~(\reffig{f:range}A).
For the roll and pitch cases, the mobility of the hand was significantly higher than the others, followed by the arm in the roll case and by the torso in the pitch case.
In terms of yaw motion, the arm significantly outperformed the remaining body parts.
On average, the hand showed to have the largest mobility ($mob_{H} = 141.6 \pm 24.9^\circ$), the arm scored second ($mob_{A} = 106.7 \pm 20.4^\circ$, $p_{HA}<0.01$) and the torso last  ($mob_{T} = 90.9 \pm 14.5^\circ$, $p_{HT}<0.01$, $p_{AT}=0.013$).

For the angular speed evaluation, we computed the difference between the average maximum and average minimum angular speed during the five fast repetitions of the movements~(\reffig{f:range}B).
The hand speed was always significantly higher.
The Hand group achieved the highest average speed ($s_{H} = 1121.3 \pm 351.9^\circ/s$), Arm group followed ($s_{A} = 372.6 \pm 111.9^\circ/s$, $p_{HA} < 0.01$), and Torso group was the slowest ($s_{T} = 285.5 \pm 72.3^\circ/s$, $p_{HT} < 0.01$, $p_{AT} = 0.019$).

\begin{figure}[th]
\begin{center}
  \includegraphics[width=\columnwidth]{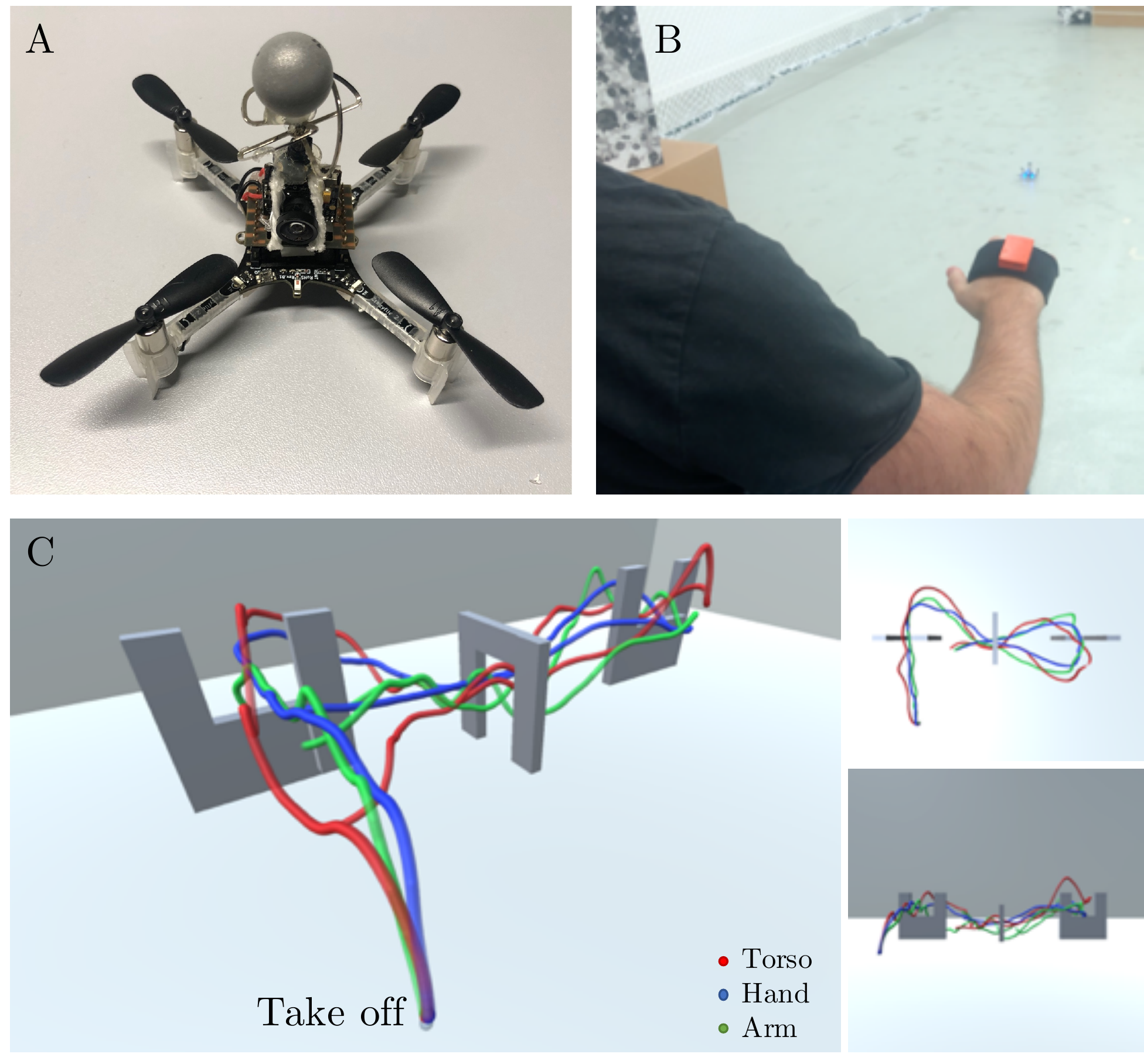}
\caption{Hardware validation scenario. (A) Crazyflie quadrotor with micro-camera for FPV video streaming. (B) Participant instrumented with IMU for teleoperation (Hand interface). (C) Segment of trajectory of the drone using the three interfaces (Unity3D rendering).}
\label{f:hw}
\end{center}
\end{figure}

\textbf{\textit{Hardware validation: }} 
we assessed the ability to transfer the teleoperation skills acquired during the simulation to a real drone through a hardware test. 
We recruited one more participant to control the flight of a quadrotor \cite{Crazyflie} using each BoMI through a path consisting of three obstacles for two runs (\reffig{f:hw}).
The quadrotor was controlled through ROS from a ground control station to mimic fixed-wing dynamics and fly at a constant speed of $0.3m/s$.
During the test, no obstacle collisions occurred.
The experiment showed that it is possible to transfer teleoperation skills for the considered set of BoMIs to the teleoperation of a real drone.

\section{Discussion}\label{sec:discussion}

In this study, we compared spontaneous body motion strategies to control drones \cite{miehlbradt_data-driven_2018, macchini_personalized_2020} with alternative motion strategies  and quantified their intuitiveness in terms of performance and user experience. Also, we found in our participants' motion a set of characteristics that correlate with these metrics and could provide a possible explanation for our results.
Here, we summarize our main findings.

Our performance test showed that basing the interface on different body segments is a condition affecting significantly the users' precision (\reffig{f:perf}).
Arm group was the least precise, with a $77\%$ higher error in hitting the waypoints.
Torso group and Hand group performed similarly in the initial stage of the experiment (\reffig{f:perf}B), but while the Torso group's average error dropped by $20\%$ with practice, the Hand group could improve their performance by $54\%$, showing a better learning capacity (\reffig{f:perf}C).
This result proves that spontaneous motion patterns provide sub-optimal interfaces that can be outperformed by choosing the motion features heuristically.
In summary, we show that humans are not able to identify the best body segment to control a robot with their body relying on their spontaneous motion.

The users' responses to our feedback survey align with the performance test.
$56\%$ of the subjects preferred the hand interface, $33\%$ the torso ad only $11\%$ the arm (\reffig{f:quest}).
Specifically, users preferring the hand considered it more sensitive and precise, while the torso was, on average, considered a more natural and immersive option. 
Interestingly, two subjects preferred the arm interface despite the lower achieved performance.
Despite being the preferred option, some participants noticed that the hand-based HRI could become tiring in the long term.
These data show that the user's feedback can be a good predictor of performance and that the quality of a wearable interface is complex to measure as it depends on a set of factors.
People consider different aspects of the teleoperation experience when choosing their preferred alternative: some focus on performance and others on the sense of presence and immersion.

By studying the participants' body motion, we provided two possible explanations for the aforementioned findings.
We first observed the motion of the three groups to control the drone.
While the Torso group and the Hand group moved in a very consistent way, i.e., controlling the same degrees of freedom of the robot with similar body gestures, for the Arm group it was not possible to identify a dominant motion pattern (\reffig{f:motion}).
Compressing human-drone motion correlation data with PCA, torso and hand users appear aggregated, while arm users presented a much higher variability ($583\%$ compared to Torso group and by $173\%$ compared to Hand group).
This intra-group inconsistency could be due to a higher difficulty for the participants to decide how to move and translate in a lower interface's effectiveness.
This observation correlates with the higher performance in the initial stage of teleoperation for the Torso group and Hand group, but does not explain the higher learning effect observed in hand users.

The second analysis is related to the motion capabilities of the various body segment (\reffig{f:range}).
The higher mobility for the hand segment is known from the literature \cite{boone_normal_1979}.
However, since a high variability in segment mobility has been observed in terms of age and gender \cite{doriot_effects_2006}, we decided to measure it on our subjects.
We showed that the hand's available mobility was on average $33\%$ larger than the arm and $56\%$ larger than the torso, and its angular speed almost three times higher than the arm and almost four times higher than the torso.
This higher dexterity might contribute to the improved performance of the hand interface, and to the perceived responsiveness when controlling the drone. 
Mobility analysis has already been proposed as a method for BoMI implementation, but its use is limited to shoulder motion for rehabilitation purposes \cite{farshchiansadegh_body_2014}.

In summary, we demonstrated that the observation of users' spontaneous motion is not sufficient to design optimal motion-based HRIs.
Our heuristic solution based on hand motion showed to achieve lower errors during teleoperation and better user feedback, while a second one based on arm motion provided poorer results.
Nonetheless, selecting features based on spontaneous motion could help to develop more immersive interfaces.
Our experiments suggest that the evaluation of additional variables, such as biomechanical (segment mobility) and behavioral (intra-group agreement) aspects could be additional predictors of the intuitiveness of an interface, and thus facilitate its design.

Finally, we extended our previous framework to generate a personalized BoMI using only an IMU, freeing ourselves from using a motion capture system. 
This extension allows naive and expert users to teleoperate a fixed-wing drone with their preferred body motion in a matter of minutes.
The method has been tested extensively in simulation and qualitatively in hardware, to assess the skill transferability to the teleoperation of a real drone (\reffig{f:hw}).

Our research opens several interesting future investigations.
First, this study is limited to a single robotic morphology and a single task: users' performance and feedback might change when controlling different robots.
Also, research shows that humans tend to employ motion synergies rather than single-limb movements to achieve functional motion~\cite{leo_synergy-based_2016}. 
Allowing the control of robots through gestures involving multiple body segments could unveil new insights into the effects of synergistic motion over HRI intuitiveness.

\section{Conclusion}\label{sec:conclusion}

In this paper, we describe new insights concerning the effectiveness of data-driven approaches for the implementation of BoMIs for telerobotics.
We show that commonly adopted methods, based on the observation of users' preferred motion patterns, might lead to sub-optimal results for the  teleoperation of mobile robots, such as fixed-wing drones.
By identifying possible alternative biomechanical and behavioral elements correlated to the HRI performance, our work represents a step towards a better understanding of the human factors affecting the efficiency of wearable telerobotic interfaces and could facilitate their implementation for future applications.

\section*{Acknowledgements}

This work was partially funded by the European Union’s Horizon 2020 research and innovation programme under grant agreement ID: 871479 AERIAL-CORE and by the Swiss National Science Foundation through grant NCCR Robotics.








\bibliographystyle{IEEEtran}
\bibliography{IEEE_abbreviated.bib,IEEE_conferences_abbreviated.bib,references}

\end{document}